\documentclass[conference]{IEEEtran}
\IEEEoverridecommandlockouts
\usepackage{cite}
\usepackage{amsmath,amssymb,amsfonts}
\usepackage{algorithmic}
\usepackage{enumitem} 
\usepackage{graphicx}
\usepackage{textcomp}
\usepackage{amsmath}
\usepackage{amssymb}
\usepackage{microtype}
\usepackage{caption}
\usepackage{xcolor}
\usepackage{float}
\usepackage{stfloats}
\usepackage[utf8]{inputenc}
\usepackage{siunitx}

\def\BibTeX{{\rm B\kern-.05em{\sc i\kern-.025em b}\kern-.08em
    T\kern-.1667em\lower.7ex\hbox{E}\kern-.125emX}}

    \usepackage{caption}
\begin{document}

\title{Agent-Based ML-LLM Fusion with Self-Optimizing Prompts for Plateau Weather Alerts\\
\thanks{This research was funded by the Sichuan Center for Education Development Research (Project No.: CJF25092).}
}

\author{\IEEEauthorblockN{1\textsuperscript{st} Shuai Yan}
\IEEEauthorblockA{\textit{Chengdu Jincheng College} \\
\textit{College of Computer and Software}\\
Chengdu, Sichuan, China  \\
yanshuai1@cdjcc.edu.cn}
\and
\IEEEauthorblockN{2\textsuperscript{nd} Yang Xu}
\IEEEauthorblockA{\textit{Chengdu Jincheng College} \\
\textit{College of Computer and Software}\\
Chengdu, Sichuan, China  \\
xuyang88@cdjcc.edu.cn}
\and
\IEEEauthorblockN{3\textsuperscript{rd}  Shan He$^{\ast}$}
\IEEEauthorblockA{\textit{Chengdu Jincheng College} \\
\textit{College of Computer and Software}\\
Chengdu, Sichuan, China  \\
heshan@cdjcc.edu.cn}

}

\maketitle

\begin{abstract}
To address insufficient contextualization, weak generalization, and poor scenario adaptation in tourism meteorological services, we propose SmartWeatherAgent---a unified three-stage architecture integrating intent recognition, hazard prediction, and reasoning-enhanced generation. The system fuses rule-based methods with large language models to parse queries at multiple granularities and employs a LightGBM model enriched with highland-specific features (e.g., wind speed abruptness rate), achieving an F1-Macro score of 0.605 with 1.60\,ms latency on high-wind, precipitation, and low-temperature events. A 12-round micro-step prompt self-optimization loop boosts the composite warning quality score \(S_{\text{final}}\) from 4.2 (B01) to 8.9 (B12, +112\%). Key improvements include a sharp rise in B08 from data source citation (6.5\,\(\rightarrow\)\,8.5), sustained high performance in B10 via physical mechanism explanation, and a peak scientific rigor score of 9.2 in B12 through explicit uncertainty statements. The system autonomously generates structured warnings that integrate causal mechanisms, spatiotemporal evolution, quantitative evidence, regulatory references, and confidence statements---enhancing professional depth, logical rigor, and scientific soundness, and advancing meteorological services toward proactive perception, explainable decision-making, and intelligent agency.
\\
\end{abstract}


\section{Introduction}
Highland tourism meteorology exhibits high dynamism, strong spatial heterogeneity, and scenario dependence, posing significant challenges to the accuracy and timeliness of existing service systems. Existing methods---such as static rule-based systems or generic large language models---commonly exhibit high response latency, poor scenario adaptation, weak context awareness, and the absence of a self-evolution mechanism, thereby failing to meet the personalized decision-making needs of tourists~\cite{b1}. To address these limitations, we propose SmartWeatherAgent---a framework that, for the first time, embeds a prompt self-adaptation mechanism into the core of an intelligent meteorological agent, inspired by the ``generation-as-reasoning'' paradigm of large language models, to establish a unified architecture comprising intent recognition, hazard prediction, and reasoning-enhanced generation~\cite{b2}. Through a closed-loop prompt refinement strategy, the system achieves: (i) fine-grained deconstruction of user query intents; (ii) efficient short-range nowcasting modeling of highland extreme weather events~\cite{b3}; and (iii) dynamic, context-aware generation of warning content---where each interaction drives prompt self-evolution.

\section{Methodology}
\subsection{Intent Recognition Module}
Combines regular expression matching with a large language model (Qwen3) to classify user intents into six categories, including ``simple inquiry,'' ``hazard alert,'' and ``family travel''~\cite{b1}. Regular expressions are applied for initial filtering, while the LLM resolves contextual ambiguities, thereby enhancing system robustness~\cite{b2}.

\subsection{Hazard Prediction Module}
Targeting three high-impact weather events prevalent in plateau regions—strong winds, precipitation, and low temperatures—this work proposes a short-term nowcasting model based on LightGBM, enhanced with plateau-specific features: Absolute wind speed difference: $\Delta W(t) = \bigl| W(t) - W(t - 1) \bigr|$; Precipitation burst indicator: $\mathbf{1}\bigl\{ P(t) > P_{95} \bigr\}$;Gust ratio: $R_g(t) = G(t) / W(t)$; Temporal encoding:$\phi_{\sin}(t) = \sin\!\left(\frac{2\pi H}{24}\right),\  \phi_{\cos}(t) = \cos\!\left(\frac{2\pi H}{24}\right)$.

Additional features include rolling statistics (e.g., 6-hour moving average of temperature), threshold-based binary features (e.g., ``diurnal temperature range $> 10^\circ$C''), and quantile-based extremeness markers (e.g., ``temperature below the 5th percentile'')~\cite{b3}. Here, $W(t)$ denotes the hourly mean wind speed, $G(t)$ the gust speed, $P(t)$ the hourly precipitation, and $H$ the local hour of day. Collectively, these features form a multidimensional input representation that supports real-time hazard prediction and provides structured grounding for the subsequent generation module.

\subsection{Prompt Self-Adaptation Module}
This module establishes a closed-loop pipeline of ``generation $\rightarrow$ evaluation $\rightarrow$ optimization'' to drive the large language model through 12 rounds of micro-step iterative refinement for prompt self-adaptation~\cite{b4}. The overall output quality is quantified by a composite score:
\begin{equation}
S_{\text{final}} = 0.35 \cdot S_{\text{semantic}} + 0.30 \cdot S_{\text{logical}} + 0.35 \cdot S_{\text{scientific}},
\end{equation}

Each component is evaluated as follows:

Assesses whether the output progressively achieves the following sequence: phenomenon description $\rightarrow$ single-cause attribution $\rightarrow$ multi-factor coupled mechanisms $\rightarrow$ regional risk differentiation $\rightarrow$ defense measures linked to underlying physical processes. Each successful transition yields a 2--3 point increment; scores of 9--10 require explicit support from climatic context and the use of nested, compound causal expressions~\cite{b5}.

 Evaluates the completeness of the reasoning chain: ``meteorological system trigger $\rightarrow$ temporal evolution $\rightarrow$ impact propagation $\rightarrow$ targeted mitigation advice.'' A single complete chain earns 7--8 points; only systems exhibiting parallel, nested reasoning chains with precise and unbroken mapping to actionable measures achieve 9--10 points.

 Computed as the sum of five dimensions: clarity of quantitative metrics, data traceability, accuracy of regulatory citations, completeness of uncertainty statements, and terminological rigor. Any missing dimension incurs a penalty; outputs containing scientific inaccuracies are capped at a maximum score of 3.

\section{Experiments}
\subsection{Model Selection and Hyperparameter Optimization}
This experiment quantifies the performance of high-altitude meteorological hazard prediction models through a two-dimensional evaluation framework: inference latency and overall classification performance (F1-Macro, denoted as $F1\text{-}M$). LightGBM was compared against Random Forest, Gradient Boosting, XGBoost, and CatBoost using an hourly meteorological observation dataset from Lhasa as the benchmark.

\paragraph{Dataset and Feature Engineering}
The experiment uses historical hourly meteorological data for Lhasa provided by the VisualCrossing platform, spanning from January 1, 2024, to May 21, 2025, comprising \num{12168} records. Core variables include temperature (\si{\celsius}), hourly precipitation (\si{\milli\meter}), mean wind speed and gust speed (\si{\kilo\meter\per\hour}), UV index, and visibility (\si{\kilo\meter}). Missing values account for less than \SI{2.1}{\percent} of the data and are imputed using linear interpolation, leveraging the temporal continuity of the time series~\cite{b4}. The dataset is split chronologically into training and test sets in a 7:3 ratio.

Significant class imbalance is observed: in the training set, normal weather accounts for \SI{91.5}{\percent}, while strong wind and precipitation constitute only \SI{1.0}{\percent} and \SI{2.5}{\percent}, respectively~\cite{b6}, highlighting the challenge posed by the low frequency of extreme events on the plateau for minority-class recognition.

To capture the rapid evolution and nonlinear dynamics of meteorological hazards, six categories of derived features are engineered:
The engineered features include: 3-hour and 6-hour rolling window statistics (mean and standard deviation); the absolute first-order difference of wind speed; the gust ratio (gust speed divided by mean wind speed); a binary indicator for diurnal temperature range exceeding \SI{10}{\celsius}; sin/cosine-encoded cyclical features for hour-of-day and month; and extreme-event indicator variables (e.g., set to 1 if hourly precipitation exceeds the 95th percentile of the training distribution).
The final feature vector has a dimensionality of 38.

\paragraph{Experimental Design}
All models undergo hyperparameter tuning via Bayesian optimization, with the objective function defined as the F1-Macro score under 5-fold time-series cross-validation (\texttt{TimeSeriesSplit}) to prevent temporal information leakage caused by random data splitting. The optimization search space covers common hyperparameters:
The following hyperparameter ranges are explored: 
\texttt{n\_estimators} $\in [100, 500]$, 
\texttt{learning\_rate} $\in [0.01, 0.3]$, 
\texttt{max\_depth} $\in [3, 10]$, 
and \texttt{subsample} $\in [0.5, 1.0]$
These ranges are aligned with official recommendations from mainstream gradient boosting frameworks and recent literature on meteorological forecasting. The results are shown in the figure1.

\begin{figure}
    \centering
    \includegraphics[width=1\linewidth]{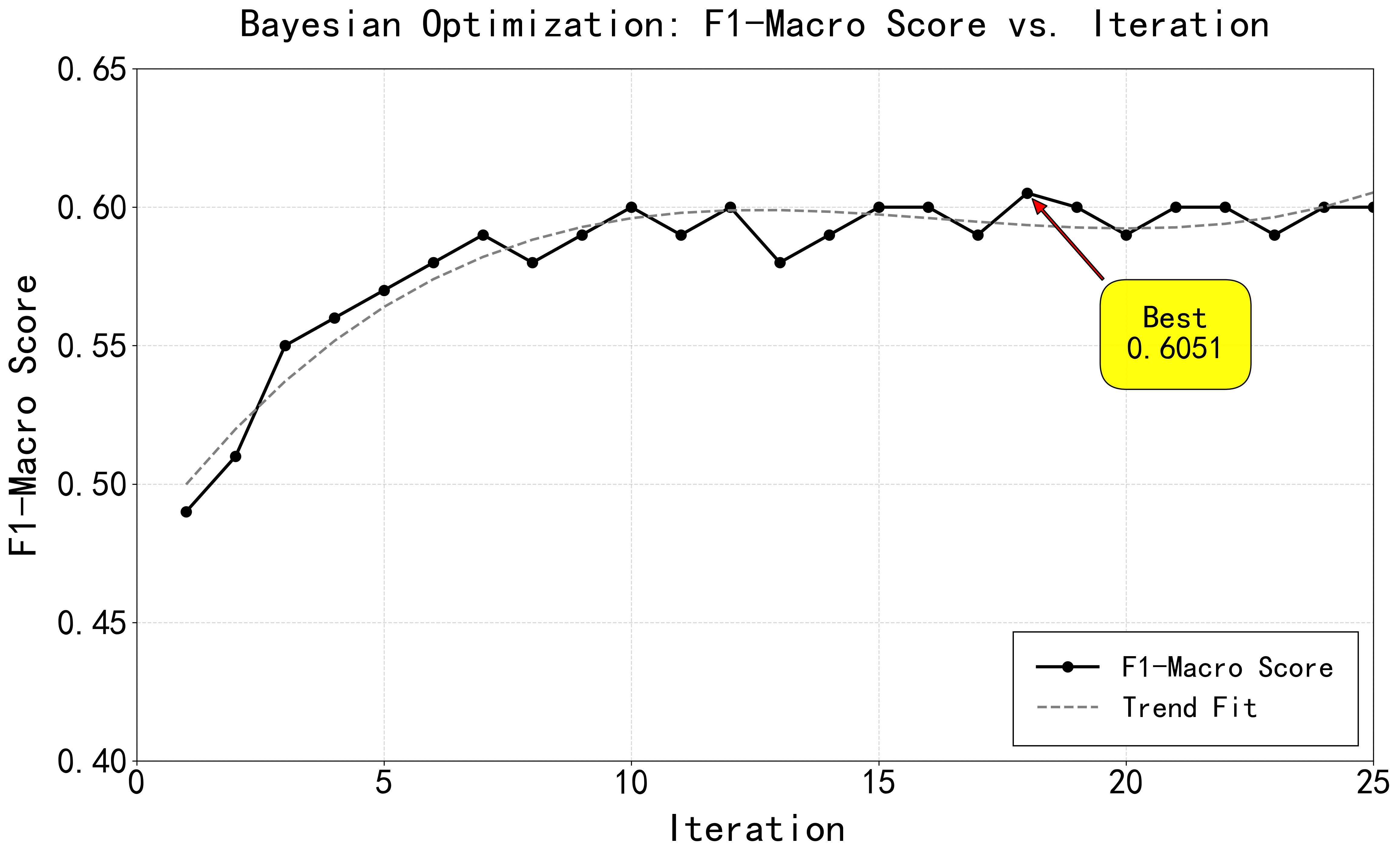}
    \caption{Visualization of Hyperparameter Optimization Process}
    \label{fig:placeholder}
\end{figure}

This study constructs hazard labels based on hourly meteorological observations from Lhasa spanning 2024–2025. Informed by the climatic characteristics of the Tibetan Plateau and the empirical data distribution, three high-impact weather events are defined as follows:
\textit{Strong wind} is defined as hourly mean wind speed $\geq \SI{20}{\kilo\meter\per\hour}$; \textit{low temperature} as hourly temperature $\leq \SI{-5}{\celsius}$; and \textit{precipitation} as hourly precipitation $> \SI{0.1}{\milli\meter}$.

Statistical analysis reveals a pronounced class imbalance in the training set: \SI{95.0}{\percent} normal weather, \SI{3.5}{\percent} low-temperature events, \SI{1.2}{\percent} strong-wind events, and only \SI{0.3}{\percent} precipitation events.

Each model undergoes 25 optimization iterations, approximating a local optimum under constrained computational budgets~\cite{b7}. To holistically evaluate the practical utility of models in plateau meteorological hazard warning, we define a composite performance score $S$ to quantify their recognition capability across critical hazard categories:
\begin{equation}
    S = 0.3 \cdot F1\text{-Macro} + 0.2 \cdot F1_g + 0.2 \cdot F1_r + 0.3 \cdot F1_c
\end{equation}
where $F1_g$, $F1_r$, and $F1_c$ denote the F1 scores for strong wind, precipitation, and low temperature, respectively.

\medskip
\noindent\textbf{Note:} The weighting scheme reflects domain considerations—although strong wind and precipitation are sparse in occurrence, they entail high risk and are thus assigned equal weight (0.2). Low-temperature events, while moderately frequent, also pose significant hazards and consequently receive the highest weight (0.3) among the specific hazard categories.

\paragraph{Experimental Results and Analysis}
As shown in Figure 2, the experimental results demonstrate that the proposed LightGBM model significantly outperforms all baseline methods in both overall warning capability ($S$) and critical hazard event recognition performance (F1 scores). As shown in Table 1, LightGBM achieves the highest composite score ($S = 0.55$), outperforming XGBoost  and Gradient Boosting .

Regarding extreme weather recognition, LightGBM attains an F1 score of $0.50$ for the precipitation category (note that this metric’s stability is limited due to sparse samples). For low-temperature events, the F1 score reaches $0.77$, markedly outperforming all baselines. Although its F1 score for strong wind events ($0.17$) is slightly lower than the best value ($0.18$ achieved by gradient boosting), the difference is minimal, indicating that LightGBM maintains strong recognition capability for high-risk, sparse events while achieving better class balance. Additionally, LightGBM’s inference latency is merely \SI{1.60}{\milli\second}, significantly outperforming XGBoost and Random Forest, thereby meeting the stringent millisecond-level real-time warning requirements of plateau tourism scenarios.

This superior performance primarily stems from LightGBM’s unique combination of histogram-based efficient feature splitting and adaptive learning strategies for sparse hazard samples, effectively mitigating the challenges posed by the dominance of normal weather samples (\SI{>91}{\percent}) and the sporadic, low-frequency nature of hazardous events (e.g., strong winds account for only \SI{1.0}{\percent}) in plateau meteorological data. Overall, the method achieves a high accuracy of \SI{97}{\percent} while maintaining relatively optimal class-balanced recognition capability ($F1\text{-macro} = 0.61$), particularly demonstrating robust performance on low-temperature and strong-wind events. By keeping inference latency within \SI{2}{\milli\second}, it provides a solid technical foundation for a high-reliability, low-latency, end-to-end warning system for meteorological hazards in plateau tourism.
\begin{figure}
    \centering
    \includegraphics[width=1\linewidth]{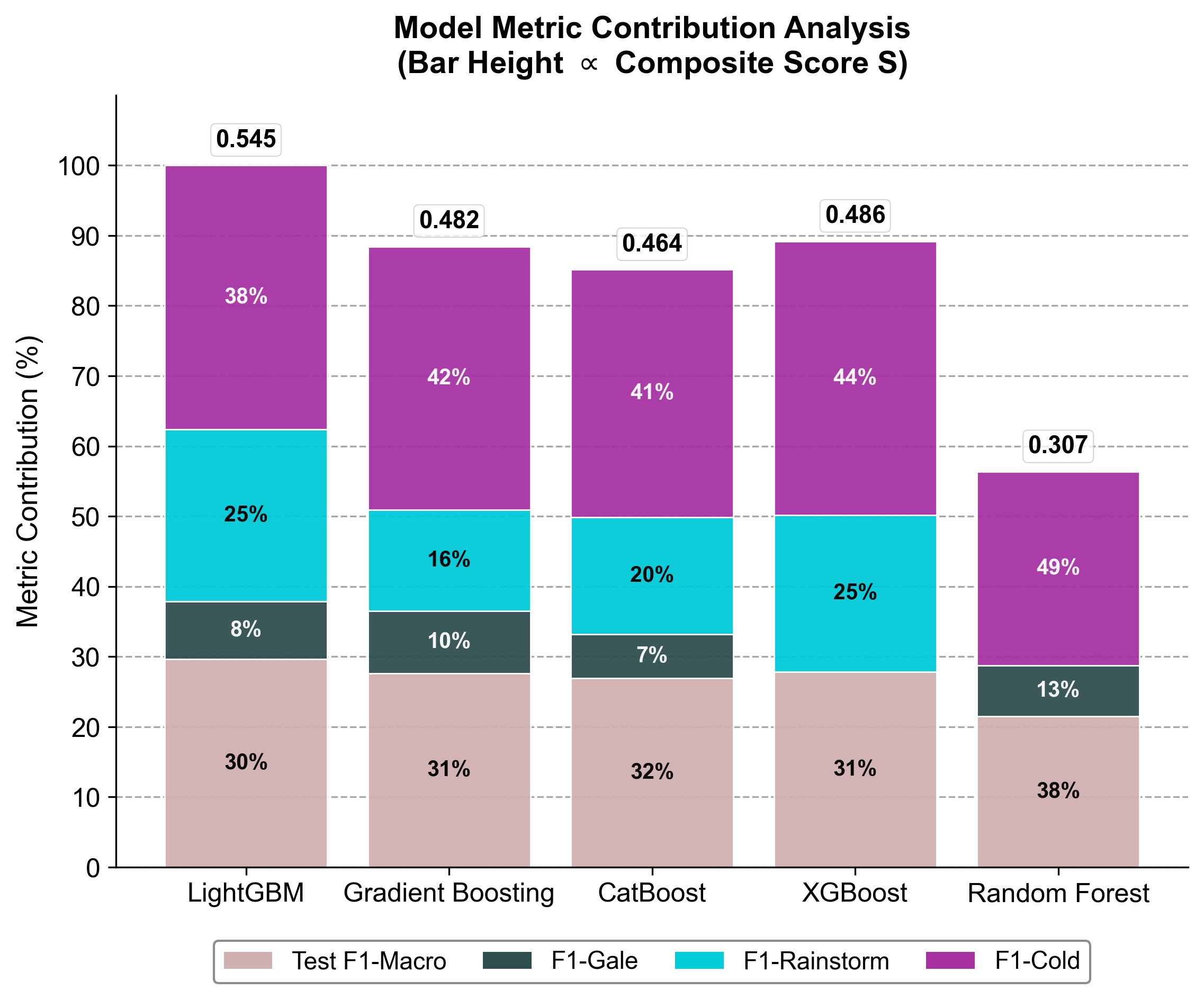}
    \caption{Performance Comparison}
    \label{fig:placeholder}
\end{figure}

\begin{table}
    \centering
    \caption{Performance Comparison of Different Models}
    \label{tab:model_performance}
    \begin{tabular}{|l|l|l|l|l|l|l|l|}\hline
          Model& F1-g&  F1-r& F1-c&F1-M&Acc&  S& Lat\\\hline
          LGBM
& 0.17
&  0.5
& 0.77
&0.61
&0.97
&  0.55
& 1.60
\\\hline
          GB
& 0.18
&  0.29
& 0.75
&0.55
&0.97
&  0.48
& 1.56
\\\hline
 CB
& 0.13
& 0.33
& 0.7
& 0.54
& 0.97
& 0.46
&1.58
\\\hline
 XGB
& 0
& 0.44
& 0.77
& 0.55
& 0.98
& 0.48
&10.32
\\\hline
 RF
& 0.14
& 0
& 0.52
& 0.41
& 0.94
& 0.31
&63.72
\\ \hline 
    \end{tabular}

\end{table}

\subsection{Self-Optimization Experiment of Meteorological Warning Prompts via Micro-Step Iteration}
To validate the efficacy of fine-grained prompt iterative optimization in approximating provincial warning standards and enhancing multi-dimensional performance, we implement a ``generation--evaluation--optimization'' closed-loop framework in a high-altitude severe convective scenario. The initial prompt, being unstructured and lacking domain-specific constraints, yielded outputs missing critical elements---namely warning levels, quantified metrics, protective actions, and regulatory justification---thereby deviating markedly from operational norms~\cite{b8}. Over the course of $12 \times 5$ rounds of micro-step refinement, we progressively introduced hierarchical constraints (``temporal $\rightarrow$ spatial $\rightarrow$ mechanistic $\rightarrow$ uncertainty'') to systematically evaluate the prompt's evolution in terms of semantic depth, logical coherence, and scientific rigor, thereby aligning with the paradigms of dynamic prompt engineering and phased objective scheduling~\cite{b9}.

\paragraph{Experimental Design}
A controlled micro-step design was adopted: the B01 baseline used informal prompts, yielding outputs that deviated significantly from the standard template, whereas the adaptive group employed a three-layer closed-loop architecture. The generation layer fed simulated meteorological features into Qwen-Max; the evaluation layer scored outputs (0.0--10.0) across semantic depth, logical coherence, and scientific rigor, with diagnostic feedback; and the optimization layer refined prompts in stages---early (B01--B04) added core warning elements (e.g., alert level, area, precautions); middle (B05--B08) emphasized quantified metrics and data traceability; and late (B09--B12) prioritized mechanistic explanations and decision-support capabilities. This staged prompt self-reconstruction integrates incremental engineering with objective scheduling, using the three-dimensional scores to compute a composite alignment metric against the \emph{Technical Regulations on the Issuance of Meteorological Disaster Warning Signals}~\cite{b10}.

\newcommand{\Sfinal}{S_{\text{final}}}
\paragraph{Experimental Results and Performance Analysis}
As shown in Figure 3, five-round averages show semantic depth rising from 2.0 (B01) to 8.5 (B12), with a +1.5 jump at B10 (``explain physical mechanisms''); logical coherence steadily increased from 7.5 to 9.0, driven by B11 (``construct temporal evolution chain''); and scientific rigor surged from 3.0 to 9.2, marked by a +2.2 gain at B08 (``cite data sources'') and a peak at B12 (``add uncertainty statements'') after a dip to 7.0 at B11. The composite score \(\Sfinal\) improved from 4.2 to 8.9 (+112\%), first exceeding 8.0 at B08 (6.5\,\(\rightarrow\)\,8.5).

During the early phase (B01--B06), \(\Sfinal\) rose gradually from 4.2 to 7.5 as essential warning elements were incorporated. In the mid phase (B07--B09), introducing quantified variables at B07 temporarily reduced semantic depth to 4.7, causing \(\Sfinal\) to dip to 6.5; B08 recovered performance via enhanced data traceability. In the late phase (B10--B12), coordinated optimization stabilized \(\Sfinal\) at 8.0--8.9, with B12 achieving a balanced high score (semantic: 8.5, logical: 9.0, scientific: 9.2). This demonstrates that integrating mechanism explanation, temporal evolution chains, and uncertainty statements effectively overcomes model limitations, unifying professional depth, structural rigor, and scientific credibility. All metrics showed standard deviations \(< 0.5\), confirming trajectory stability and reproducibility.
\begin{figure}
    \centering
    \includegraphics[width=1\linewidth]{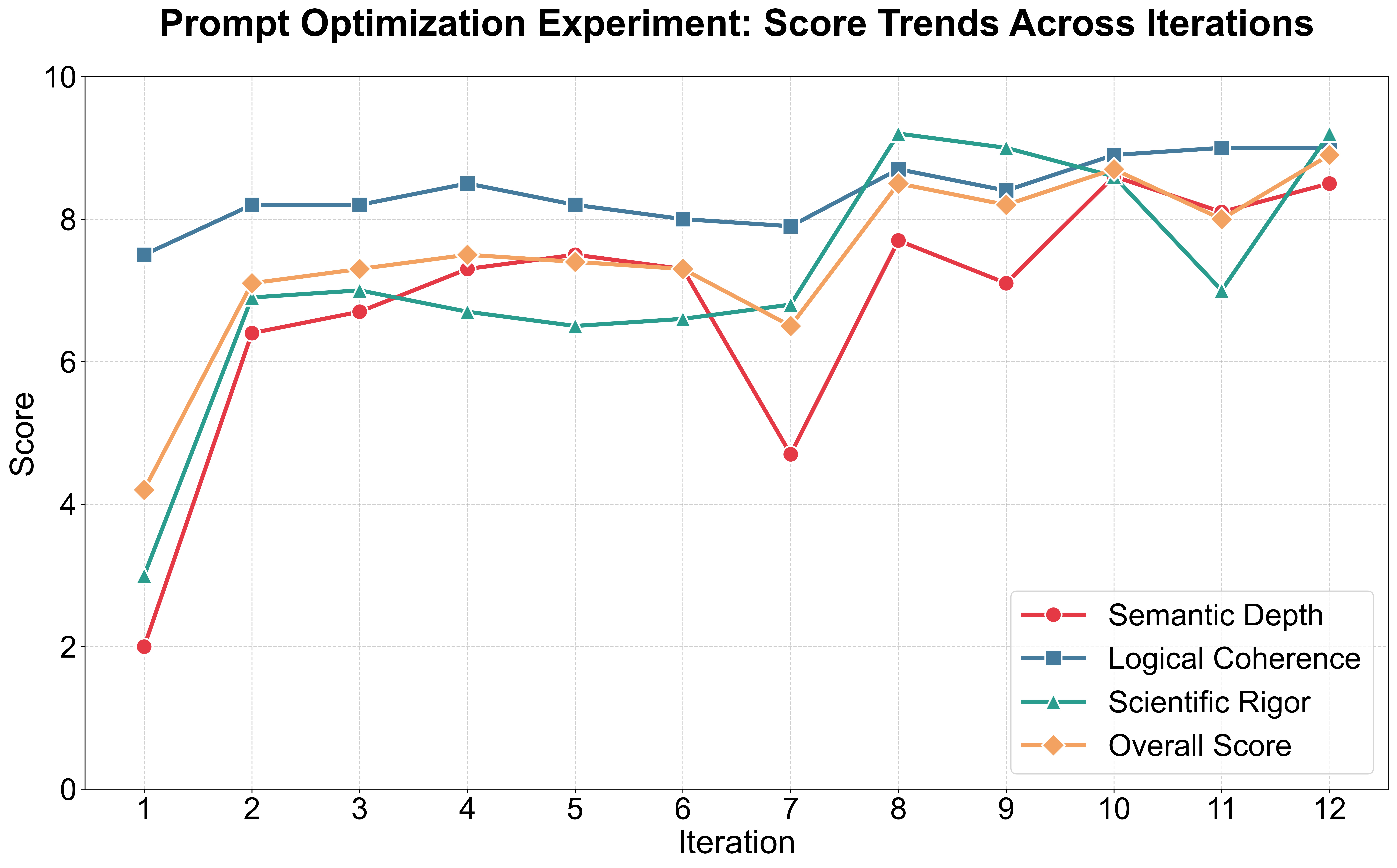}
    \caption{prompt optimization scores}
    \label{fig:placeholder}
\end{figure}

\section{Discussion and Limitations}
Although the proposed method performs well in intelligent weather warning tasks for highland tourism cities, several limitations remain. First, the training and validation data are limited to Lhasa and are not representative of other highland cities, limiting the model’s geographic generalization capability. Second, although the prompt optimization over 12 rounds achieves end-to-end autonomous refinement---e.g., B08 automatically incorporating data provenance and B12 generating statements of uncertainty---the evolutionary process remains constrained by predefined evaluation dimensions and a staged framework. It lacks the ability to openly perceive emerging warning needs and to structurally self-reconfigure, thereby limiting the system’s sustained adaptability in dynamic, complex scenarios. Third, the system is trained solely on historical observational data and has not been integrated with real-time operational meteorological data streams; consequently, its robustness and real-world effectiveness under challenging conditions---such as communication outages, sensor noise, or extremely rare events---require validation through field deployment.

\section{Conclusion}
This paper presents SmartWeatherAgent, an end-to-end intelligent weather service framework for short-range nowcasting in highland tourism, integrating intent recognition, a lightweight high-impact weather predictor, and LLM-based generation. Its three-stage, feedback-driven architecture---generation, evaluation, and optimization---iteratively refines warning messages to enhance professionalism, structural integrity, and alignment with meteorological standards. The highland-optimized predictor accurately detects key hazards  while remaining sensitive to sparse precipitation, ensuring reliability and timeliness in complex terrain. The system’s modular, self-evolving, and scenario-adaptive design makes it applicable beyond tourism---to broader public safety and emergency response contexts requiring real-time awareness. Future work will integrate real-time meteorological data, support multimodal interaction, and develop a meta-prompt-driven self-reflective optimizer to boost generalization, interaction naturalness, and output credibility

\vspace{12pt}

\end{document}